\pdfoutput=1

\documentclass[11pt]{article}

\usepackage[preprint]{acl}
\usepackage{subcaption}
\usepackage{times}
\usepackage{latexsym}

\usepackage[T1]{fontenc}

\usepackage[utf8]{inputenc}

\usepackage{microtype}

\usepackage{inconsolata}

\usepackage{graphicx}
\usepackage{amsmath}
\usepackage{amssymb}
\usepackage{multirow}
\usepackage{booktabs}
\usepackage{enumitem}
\usepackage{pifont}

\usepackage{fontawesome}
\usepackage{xspace}
\newcommand{\huggingface}{\raisebox{-1.5pt}{\includegraphics[height=1.05em]{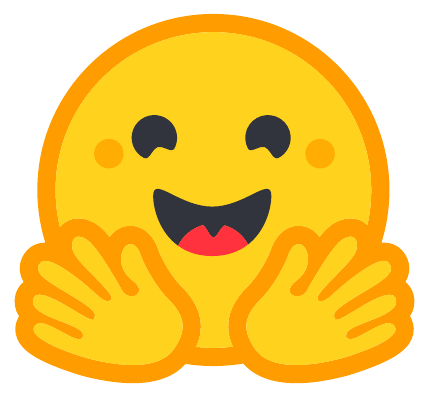}}\xspace}
\newcommand{\github}{\raisebox{-1.5pt}{\includegraphics[height=1.05em]{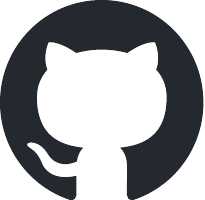}}\xspace}

\newcommand{\xmark}{\small{\usym{2613}}} 
\usepackage{utfsym}
\usepackage[table]{xcolor}

\usepackage{listings}
\usepackage{color}

\definecolor{dkgreen}{rgb}{0,0.6,0}
\definecolor{gray}{rgb}{0.5,0.5,0.5}
\definecolor{mauve}{rgb}{0.58,0,0.82}

\title{
Gradient-Attention Guided Dual-Masking Synergetic Framework for Robust Text-based Person Retrieval}

\author{Tianlu Zheng$^\text{\ding{170}}$\footnotemark[1], 
Yifan Zhang$^{\text{\ding{105}}}$\footnotemark[1], 
Xiang An$^{\text{\ding{171}}}$, 
Ziyong Feng$^{\text{\ding{171}}}$ \\ 
\textbf{Kaicheng Yang}$^{\text{\ding{171}}}$\footnotemark[2], 
\textbf{Qichuan Ding}$^{\text{\ding{170}}}$\footnotemark[2]\\
$^{\text{\ding{170}}}$Northeastern University
$^{\text{\ding{105}}}$South China University of Technology $^{\text{\ding{171}}}$DeepGlint \\
\texttt\small{\small 2302190@stu.neu.edu.cn, \small dingqichuan@mail.neu.edu.cn, \small kaichengyang@deepglint.com} \\ 
\small{\github Code:  \url{https://github.com/Multimodal-Representation-Learning-MRL/GA-DMS}} \\ 
\small{\huggingface Data:  \url{https://huggingface.co/datasets/Kaichengalex/WebPerson-5M}}}

\begin{document}
\maketitle
\renewcommand{\thefootnote}{\fnsymbol{footnote}}
\footnotetext[1]{Equal contribution.}
\footnotetext[2]{Corresponding author.}

\begin{abstract}
Although Contrastive Language-Image Pre-training (CLIP) exhibits strong performance across diverse vision tasks, its application to person representation learning faces two critical challenges: \textit{(i)} the scarcity of large-scale annotated vision-language data focused on person-centric images, and \textit{(ii)} the inherent limitations of global contrastive learning, which struggles to maintain discriminative local features crucial for fine-grained matching while remaining vulnerable to noisy text tokens. This work advances CLIP for person representation learning through synergistic improvements in data curation and model architecture. First, we develop a noise-resistant data construction pipeline that leverages the in-context learning capabilities of MLLMs to automatically filter and caption web-sourced images. This yields \textbf{WebPerson}, a large-scale dataset of 5M high-quality person-centric image-text pairs. Second, we introduce the \textbf{GA-DMS} (\textbf{G}radient-\textbf{A}ttention Guided \textbf{D}ual-\textbf{M}asking \textbf{S}ynergetic) framework, which improves cross-modal alignment by adaptively masking noisy textual tokens based on the gradient-attention similarity score. Additionally, we incorporate masked token prediction objectives that compel the model to predict informative text tokens, enhancing fine-grained semantic representation learning. Extensive experiments show that GA-DMS achieves state-of-the-art performance across multiple benchmarks. 

\end{abstract}
\section{Introduction}
The rapid advancement of large-scale vision-language pre-training \cite{chen2023vlp, alip, rwkvclip, wu2023grounded} has been driven by the unprecedented availability of web-sourced image-text pairs. As a milestone in vision-language representation learning, Contrastive Language–Image Pre-training (CLIP) \cite{radford2021learning} employs dual encoders for visual and textual modalities and leverages a contrastive loss mechanism \cite{wang2021understanding} to learn joint representations. Trained on 400 million noisy web-curated image-text pairs, CLIP exhibits strong zero-shot generalization and has been widely adopted for tasks including image classification \cite{abdelfattah2023cdul, peng2023sgva, unicom}, retrieval \cite{sain2023clip, shao2023unified,decoupled}, and grounding \cite{xiao2023clip,mlcd,rice}. However, CLIP shows suboptimal performance in text-based person retrieval, as evidenced by recent studies \cite{shao2023unified, yan2023clip, li2023clip, han2024clip, zhao2025gradeclip}.

\begin{figure}[t!]
\begin{subfigure}[b]{\linewidth}
\centering
\includegraphics[width=1.0\linewidth]{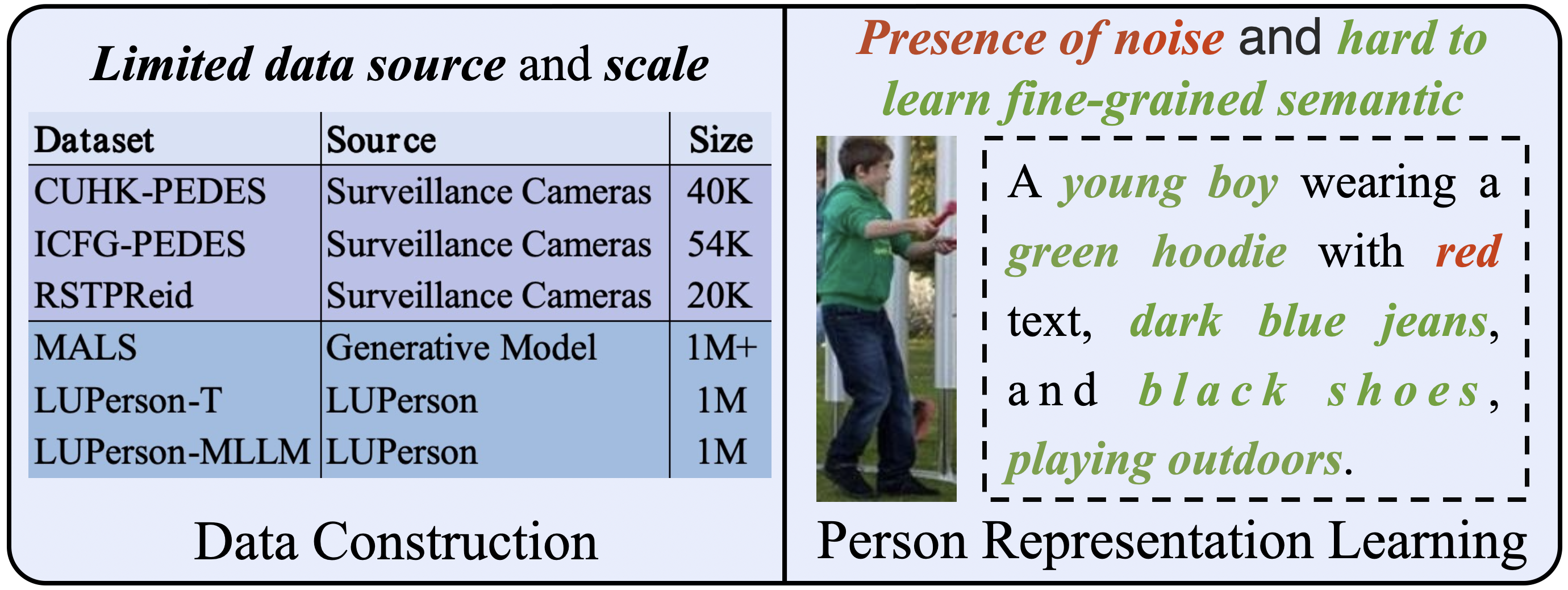}
\vspace{-6mm}
\subcaption{\small Current work exhibits several deficiencies}
\end{subfigure}
\hfill
\hfill
\begin{subfigure}[b]{\linewidth}
\centering
\includegraphics[width=1.0\linewidth]{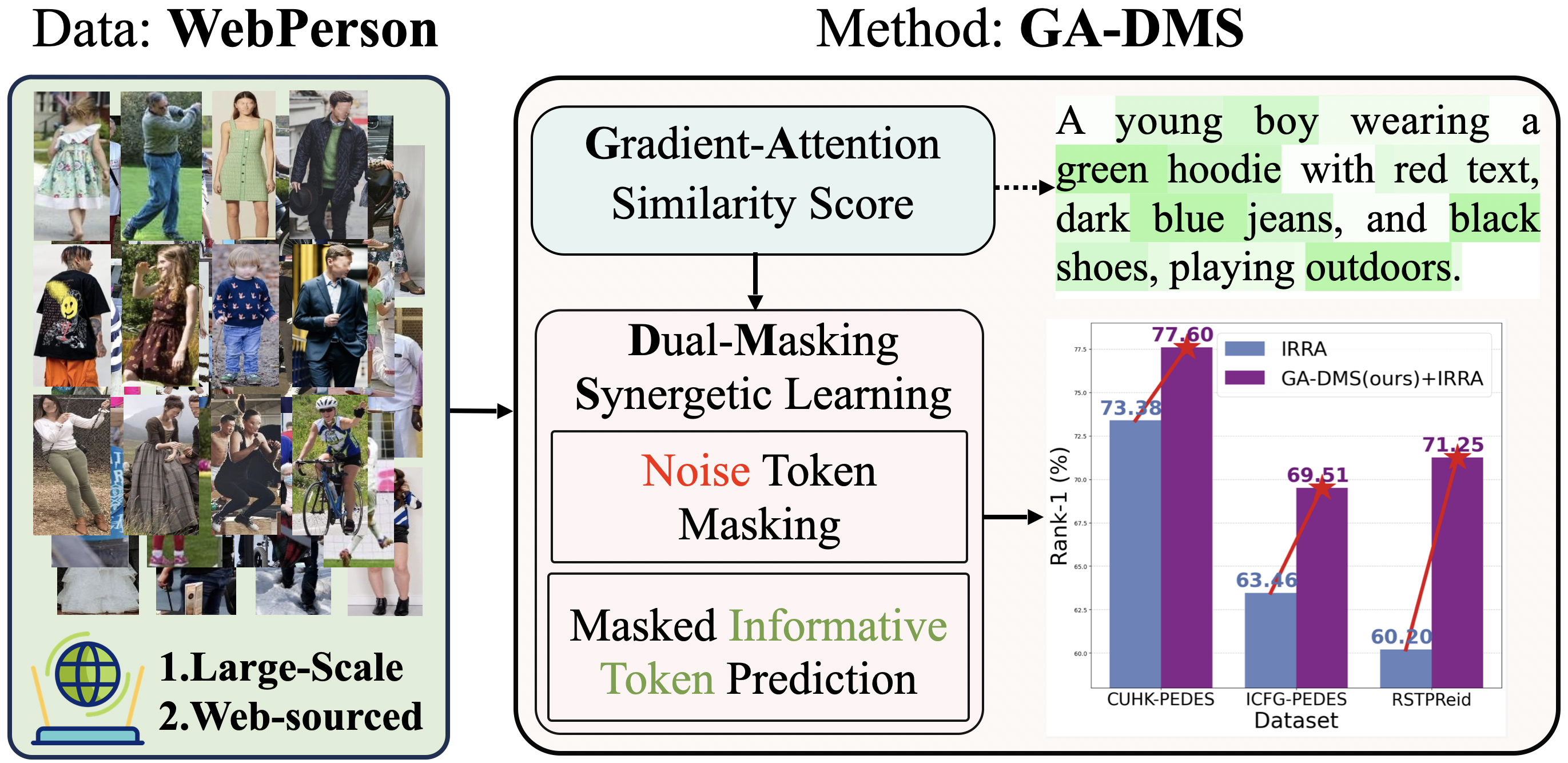}
\vspace{-6mm}
\subcaption{\small Our method for robust person representation learning}
\end{subfigure}
\vspace{-7mm}
\caption{Current human-centric datasets are limited in diversity and scale, complicating model training due to noise interference and hindering the effective learning of fine-grained semantics.}
\vspace{-5mm}
\label{fig:motivation}
\end{figure}

CLIP's suboptimal performance in text-based person retrieval stems from two key limitations. First, the scarcity and noise levels in person-centric image-text data pose significant challenges. Existing datasets such as CUHK-PEDES \cite{li2017person}, ICFG-PEDES \cite{ding2021semantically}, and RSTPReid \cite{zhu2021dssl} are constrained in scale due to their reliance on extensive manual annotations. Although large-scale person-centric datasets like LUPerson \cite{fu2021unsupervised} comprise approximately 200K identities and 4 million images, they lack corresponding textual descriptions. Recent efforts~\cite{tan2024harnessing} have employed Multimodal Large Language Models (MLLMs) to address data scarcity by generating synthetic captions. However, these automatically produced annotations frequently contain inaccuracies and semantic misalignments, thereby introducing noise into the training process and requiring the implementation of corrective strategies \cite{zhao2024genixer}. Second, CLIP's global contrastive learning paradigm fails to effectively capture fine-grained visual semantics crucial for distinguishing similar individuals~\cite{yan2023clip, liu2024distilling}. This is particularly problematic as person retrieval often relies on localized attributes (\textit{e.g.}, clothing patterns or accessories) that require precise visual-semantic alignment.

In this work, we advance CLIP for person representation learning through synergistic improvements in data curation and model architecture~(Fig.\ref{fig:motivation}). We initially introduce \textbf{WebPerson}, a large-scale person-centric dataset consisting of 5 million high-quality text-image pairs derived from web-sourced images. After that, we propose the \textbf{GA-DMS}~(\textbf{G}radient-\textbf{A}ttention Guided \textbf{D}ual-\textbf{M}asking \textbf{S}ynergetic) framework, which enhances cross-modal alignment by masking noisy textual tokens based on a gradient-attention similarity score. Meanwhile, we incorporate masked token prediction objectives to enforce the model to predict informative text tokens, thereby enhancing fine-grained semantic representation learning. Extensive experiments demonstrate that GA-DMS establishes new state-of-the-art performance across multiple benchmarks. The main contributions of this paper are summarized as follows:
\begin{itemize}[noitemsep,topsep=0pt]
\item We \textbf{design a novel person-centric data construction pipeline} that automatically filters and annotates web-sourced images, yielding the \textbf{WebPerson} dataset with 5 million high-quality image-text pairs.

\item We \textbf{propose the GA-DMS}~(\textbf{G}radient-\textbf{A}ttention Guided \textbf{D}ual-\textbf{M}asking \textbf{S}ynergetic) framework to improve cross-modal alignment through gradient-attention guided noisy text token masking while enhancing fine-grained visual-semantic correspondence via masked informative token prediction objectives.

\item We \textbf{conduct comprehensive experiments and demonstrate that GA-DMS achieves new state-of-the-art performance} across multiple text-based person retrieval datasets.
\end{itemize}

\section{Related Works}

\subsection{Person Representation Learning}
Early approaches to text-based person retrieval typically employ separate vision and language encoders with custom alignment losses~\cite{zheng2020dual, si2018dual}. These methods often exhibit suboptimal modality alignment and require extensive manual annotation. The introduction of CLIP~\cite{radford2021learning} establishes a unified vision-language embedding space, significantly advancing cross-modal matching. Recent works extend CLIP with specialized modules for text-based person retrieval. IRRA~\cite{jiang2023cross} merges visual cues into textual tokens via a cross-modal transformer and aligns global similarity distributions. MDRL~\cite{yang2024multi} designs a cross-modality global feature learning architecture to learn the global features from the two modalities and meet the demand of the task. UniPT~\cite{shao2023unified} utilizes a simple vision-and-language pre-training framework to explicitly align the feature space of the image and text modalities during pre-training. However, these approaches largely ignore data noise, which critically influences cross-modal alignment in feature space. RDE~\cite{qin2024noisy} mitigates the adverse impact of noisy through the proposed confident consensus division and novel triplet alignment loss. ProPOT~\cite{yan2024prototypical} transforms the identity-level matching problem into a prototype learning problem, aiming to learn identity-enriched prototypes. However, prototype aggregation compromises fine-grained semantic learning.

\subsection{Person-centric Dataset}
High-quality image-text paired datasets are essential for learning discriminative person representations. However, existing manually annotated datasets (e.g., CUHK-PEDES \cite{li2017person}, ICFG-PEDES \cite{ding2021semantically}, RSTPReid \cite{zhu2021dssl}) face severe scalability limitations due to labor-intensive annotation processes. This scalability bottleneck ultimately constrains models' capacity to acquire diverse semantic information and learn discriminative features. Recent efforts to mitigate this issue focus on constructing large-scale datasets, such as LUPerson \cite{fu2021unsupervised}, LUPerson-T \cite{shao2023unified}, LUPerson-MLLM \cite{tan2024harnessing}, and SYNTH-PEDES \cite{zuo2024plip} demonstrate that increased data volume improves general pedestrian representation learning. Nevertheless, these datasets primarily derive from video sources, inheriting inherent scalability constraints from computationally intensive video processing pipelines. The success of multimodal large language models in cross-modal understanding \cite{yu2024capsfusion} has inspired their application to synthetic data generation. For instance, LUPerson-MLLM \cite{tan2024harnessing} employs template-guided MLLMs to generate diverse textual descriptions, significantly enhancing text-to-image ReID performance. However, this approach remains limited by its dependence on existing LUPerson image collections.

\section{WebPerson Dataset}

\subsection{Person-Centric Image Filtering}
\begin{figure}[t]
\centerline{\includegraphics[width=\linewidth]{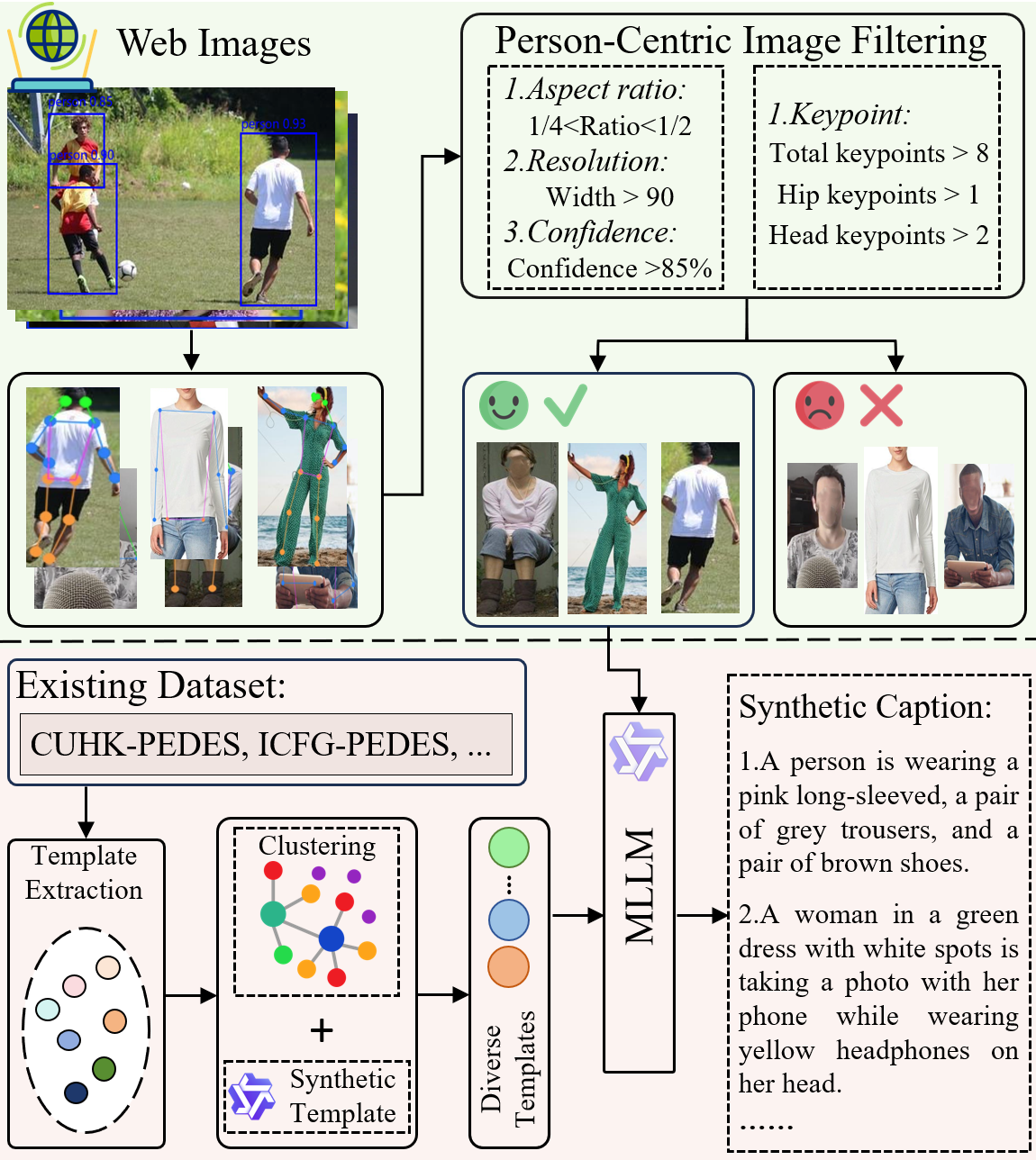}}
  \vspace{-2mm}
  \caption{The details of person-centric image filtering and synthetic caption generation pipeline for constructing our WebPerson dataset.}
  \label{fig:pipeline}
  \vspace{-5mm}
\end{figure}

In this study, we utilize the COYO700M dataset~\cite{kakaobrain2022coyo-700m}, a large-scale dataset that contains 747M image-text pairs collected from CommonCrawl, as our web-crawled images source. To filter high-quality person-centric images, we initially deploy the YOLOv11 model \cite{yolo11_ultralytics} to detect humans and extract bounding box coordinates. The specific workflow is illustrated in Fig.~\ref{fig:pipeline}, where images are retained based on the following criteria: \textit{(i)} shorter dimension exceeds 90 pixels, \textit{(ii)} aspect ratio between 1:2 and 1:4, and \textit{(iii)} human detection confidence above 85\%. Subsequently, YOLOv11-Pose \cite{yolo11_ultralytics} verifies pose integrity by requiring: \textit{(i)} visibility of at least eight keypoints, \textit{(ii)} presence of at least one hip keypoints and two head keypoints. This process yields 5 million high-quality human-centric images filtered from the COYO700M dataset.

\subsection{Synthetic Caption Generation}

Following the selection of 5 million high-quality human-centric images, we develop a synthetic caption generation pipeline to create diverse and precise textual descriptions. Our approach transforms existing captions from CUHK-PEDES~\cite{li2017person}, ICFG-PEDES~\cite{ding2021semantically}, and RSTPReid \cite{zhu2021dssl} into structured templates using Qwen2.5-72B-Instruct \cite{yang2024qwen2}. The model systematically replaces fine-grained attributes (\textit{e.g.}, black jacket, ponytail) with standardized placeholders (\textit{e.g.}, \texttt{[colored top]}, \texttt{[hairstyle]}). 

\begin{figure*}[t]
    \centering
    \includegraphics[width=\linewidth]{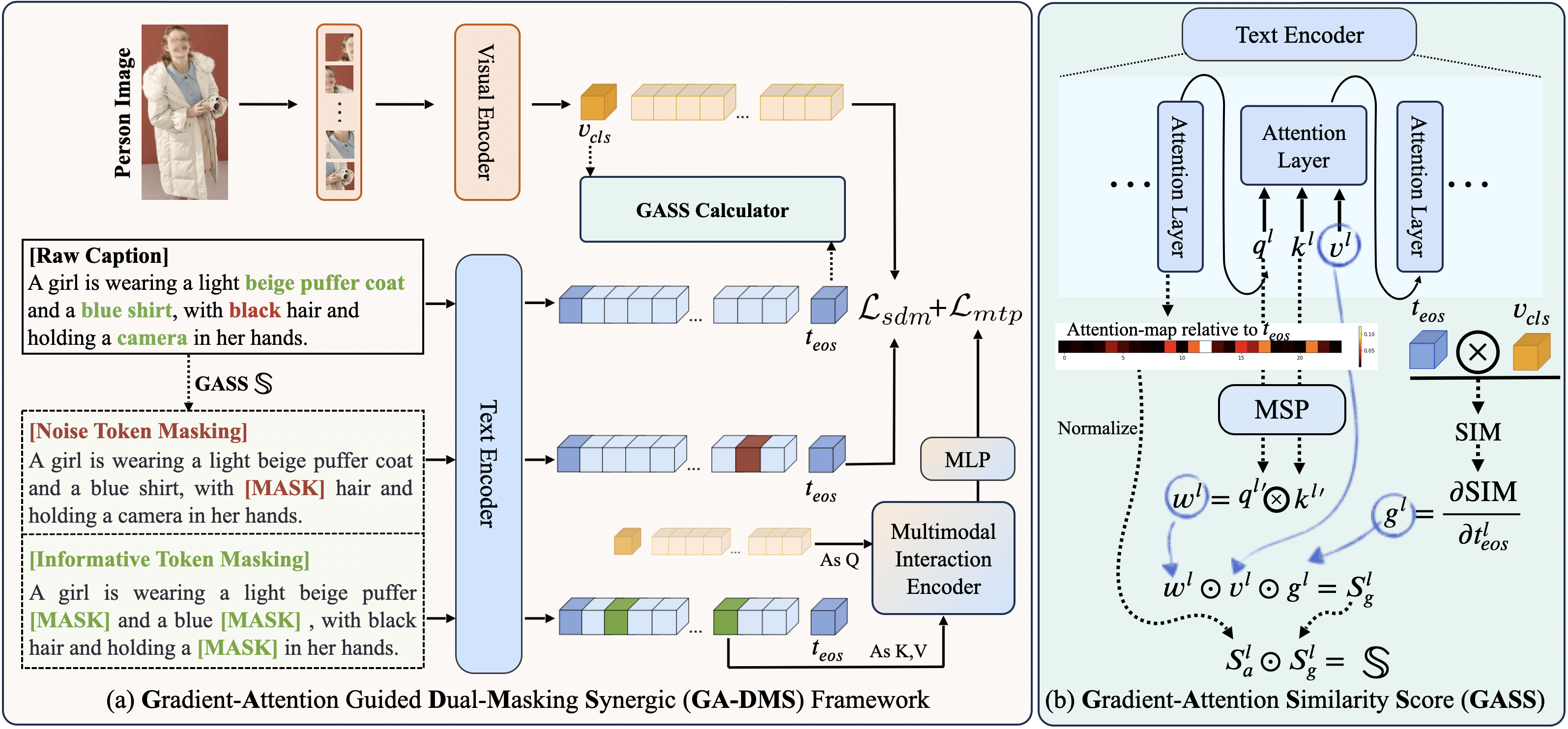}
    \vspace{-7mm}
    \caption{Overview of our proposed method. (a) The architecture of our proposed Gradient-Attention Guided Dual-Masking Synergic~(GA-DMS) framework. (b) The details of the Gradient-Attention Similarity Score~(GASS).}
    \label{fig:method}
    \vspace{-5mm}
\end{figure*}

To reduce redundancy and cluster semantically similar templates, inspired by the previous works~\cite{clipcid,realsyn}, we employ the OPENCLIP ViT-bigG/14~\cite{Radford2021LearningTV} to extract text embeddings of the template texts, then we utilize the standard $k$-means algorithm to partition all the templates embeddings into $k$ distinct clusters based on the nearest neighbor criterion. Within each cluster, we select the most representative template (highest cosine similarity to the centroid) along with five randomly sampled templates. To further enhance template diversity, we employ Qwen2.5-72B-Instruct \cite{yang2024qwen2} to synthesize new templates from this refined set. All generated templates undergo rigorous review to eliminate biases and stereotypes, yielding a curated collection of one thousand high-quality templates. To generate diverse, high-quality captions, we leverage the in-context learning capabilities of MLLMs~\cite{li2025taco,unime}. Specifically, we randomly assign a template to each image and use Qwen2.5-VL-7B-Instruct and Qwen2.5-VL-32B-Instruct \cite{bai2025qwen2} to produce captions that follow the given template. We adopt vLLM~\cite{kwon2023efficient} to accelerate large-scale inference. The details of the prompt are provided in the Appendix ~\ref{Appendix prompt}.

\section{GA-DMS Framework}
This section presents our \textbf{GA-DMS}~(\textbf{G}radient-\textbf{A}ttention Guided \textbf{D}ual-\textbf{M}asking \textbf{S}ynergetic) framework~(Fig.~\ref{fig:method}). In Sec.~\ref{section:importance}, we introduce the Gradient-Attention Similarity Score, which dynamically differentiates noise tokens and informative tokens during the training process. In Sec.~\ref{section:learning}, we present the dual-masking synergetic learning details and the training objective.

\subsection{Gradient-Attention Similarity Score}
\label{section:importance}
Existing interpretability research~\cite{selvaraju2017grad} on CLIP-based models has shown that intermediate layer gradients retain fine-grained image-text alignment information. Motivated by prior work~\cite{zhao2025gradeclip}, we introduce a gradient-attention similarity score $\mathbb{S}$ that quantifies each textual token’s contribution to the image–text alignment. We denote the embeddings of the text tokens and image tokens as $T$ and $V$. We first calculate the global cosine similarity $\mathrm{SIM} = T_{eos} V_{cls}^{\mathsf{T}}$. The gradient importance for the $l$-th transformer layer's text token $T_{eos}^l$ is then derived as $g^l= \tfrac{\partial\,\mathrm{SIM}}{\partial\,T_{eos}^l}$.

To capture fine-grained semantics, we integrate a Multi-Scale Pooling (MSP) layer within the transformer architecture. The MSP layer aggregates local contexts at multiple scales $c \in \mathcal{C}$ through average pooling of adjacent $c$ tokens, followed by bilinear interpolation to restore original dimensions. This process yields features enriched with multi-scale local information. The spatial importance $w^{l}$ for each transformer layer $l$ is then computed as:
\begin{equation}
\small
  w^l = \Phi(\mathrm{MSP}(q^l_{eos})\ \mathrm{MSP}(k^l)^{\mathsf{T}})
\end{equation}
where $\Phi$ is the normalization function, $q^l_{eos}$ is the \textit{query} embedding for the \texttt{[eos]} token at layer $l$, $k^l$ represent the \textit{key} embedding at layer $l$. The gradient-based score $S_g^{l}$ of the $l$-th transformer layer is defined as:
\begin{equation}
\small
S_g^l = g^l * w^l * v^l
\end{equation}
where $v^l$ is the \textit{value} embedding at layer $l$.

Simultaneously, we compute attention-based semantic scores $S_a^l$ for each token based on the attention maps $\mathcal{M}^l$ from the $l$-th transformer layer. We denote the attention score for the \texttt{[eos]} token as $\mathcal{W}^l$, the attention-based semantic score $S_a^l$ is computed as:
\begin{equation}
\small
S_a^l=\tfrac{\mathcal{W}^l}{\sum_{j=1}^{N} \mathcal{W}^l_{j}}
\end{equation}
The final gradient-attention similarity score $\mathbb{S}$ for the effective textual tokens is defined as:
\begin{equation}
\small
\mathbb{S} = ReLU(\tfrac{1}{L} \sum_{l \in L} S_g^l * S_a^l)
\end{equation}
where $L$ represents the number the final $L$ layers of the transformer, $\mathbb{S} \in \mathbb{R}^{B \times N}$, and $N$ is the number of tokens. This score integrates information from both gradients and attention maps to weight text tokens for masking probability computation.

\subsection{Dual-Masking Synergetic Learning}
\label{section:learning}

\subsubsection{Noise Token Masking}
\label{section:noise}
While Multimodal Large Language Models (MLLMs) inevitably introduce noise during large-scale data generation due to inherent hallucination effects. To mitigate this issue, we employ a noise token masking strategy to reduce the influence of noise tokens based on the gradient-attention similarity score $\mathbb{S}$. We calculate the masking probability for the $i$-th text token $T_i$ as:
\begin{equation}
\small
p(T_i) = \frac{\alpha_n}{1 + e^{-\lambda[(1-s_i) - \gamma ]}} 
\label{equation:5}
\end{equation}
where $s_i \in \mathbb{S}$ is the gradient-attention similarity score for the $i$-th token, $\alpha_n$ is a hyperparameter to set the upper limit of the masking probability for noise tokens. $\lambda$ and $\gamma$ respectively modulate the slope and midpoint of the probability distribution, thereby sharpening the differentiation between noisy and semantically relevant tokens. During training, we dynamically mask textual tokens using $\texttt{[mask]}$ according to these computed probabilities.

Given the embeddings of image-text pairs $\{(v_{\text{cls}}^i, t^i_{\text{eos}})\}_{i=1}^{B} $, we define the ground-truth matching distribution as $q_{i,j}$ and compute the predicted distribution as:
\begin{equation}
\small
p_{i,j} = \frac{\exp(\text{sim}(v^i_{\text{cls}}, t^j_{\text{eos}})/\tau)}{\sum_{b=1}^{B} \exp(\text{sim}(v^i_{\text{cls}}, t^b_{\text{eos}})/\tau)}
\end{equation}
where $\tau$ is a temperature parameter. Following~\cite{jiang2023cross}, we adopt the Similarity Distribution Matching (SDM) loss to align the distribution.
The $\mathcal{L}_{\text{i2t}}$ is defined as:
\begin{equation}
\small
\mathcal{L}_{\text{i2t}} = \frac{1}{B} \sum_{i=1}^{B} \sum_{j=1}^{B} p_{i,j} \log \left( \frac{p_{i,j}}{q_{i,j} + \varepsilon} \right)
\end{equation}
where $\varepsilon$ is a small number to avoid numerical problems. We compute a symmetric loss $ \mathcal{L}_{\text{t2i}} $ by swapping $\{(v_{\text{cls}}^i, t^i_{\text{eos}})\}$, and the SDM loss is:
\begin{equation}
\small
\mathcal{L}_{\text{sdm}} = \mathcal{L}_{\text{i2t}} + \mathcal{L}_{\text{t2i}}
\end{equation}

\subsubsection{Masked Informative Token Prediction}
\label{section:semantic}
To improve fine-grained semantic representation, we selectively mask tokens with strong image-semantic correlations and introduce a masked token prediction task to enhance local semantic learning. Similar to the Equation~\ref{equation:5}, the masking probability for the informative text tokens is defined as:
\begin{equation}
\small
p(T_i)=\frac{\alpha_i}{1 + e^{-\lambda[s_i - \gamma ]}} 
\end{equation}
where $\alpha_i$ bounds the maximum masking probability for informative tokens. For effective fine-grained visual-textual fusion during token prediction, we integrate a cross-modal interaction module~\cite{jiang2023cross} as a decoder. This module consists of a multi-head cross-attention followed by four Transformer layers to align modalities in a shared embedding space. A final MLP layer predicts original tokens from the fused representations. Given hidden states ${h_i^m, i \in \mathcal{M}}$ and $\mathcal{M}$ denotes the masked text token set, the distribution of the output token is $\mathbf{m}_i = \text{MLP}(h^m_i)$. The Masked Token Prediction (MTP) loss $\mathcal{L}_{mtp}$ is defined as:
\begin{equation}
\small
\mathcal{L}_{mtp}=-\frac{1}{|\mathcal{M}||\mathcal{V}|}\sum_{i\in\mathcal{M}}\sum_{j\in|\mathcal{V}|}y_j^i\log\frac{\exp(\mathbf{m}_j^i)}{\sum_{k=1}^{|\mathcal{V}|}\exp(\mathbf{m}_k^i)},
\end{equation}
where $|\mathcal{V}|$ is the size of vocabulary $\mathcal{V}$, and $y_j$ is a one-hot vocabulary distribution. Finally, the total loss $\mathcal{L}$ is define as:
\begin{equation}
\small
\mathcal{L} = \mathcal{L}_{sdm} + \beta \mathcal{L}_{mtp}
\end{equation}
where $\beta$ is a loss weight.

\section{Experiments}
\noindent{\bf Implementation Details.} Consistent with previous works~\cite{tan2024harnessing,jiang2023cross}, we utilize the CLIP ViT-B/16 model as our backbone. Following IRRA~\cite{jiang2023cross}, we incorporate a randomly initialized multimodal interaction encoder to facilitate masked token prediction. 
Our implementation processes 384×128 resolution images with a maximum length of $N=77$ text sequences. We employ Adam~\cite{kingma2014adam} as the optimizer, initialized with a learning rate of $1e-4$ and a weight decay of $4e-5$. The parameters $\beta_1$ and $\beta_2$ are set to 0.9 and 0.999, respectively. The temperature parameter $\tau$ in SDM loss is set to 0.02. We train GA-DMS for 30 epochs with a batch size of 512 on 8 NVIDIA A100~(80G) GPUs. For generating synthetic templates and captions, we utilize Qwen2.5-72B-Instruct~\cite{yang2024qwen2}, Qwen2.5-VL-7B-Instruct, and Qwen2.5-VL-32B-Instruct~\cite{bai2025qwen2}. Additionally, vLLM~\cite{kwon2023efficient} is leveraged to accelerate large-scale inference. Please refer to the Appendix ~\ref{Appendix hyper} for more detailed hyperparameters.

\begin{table*}[t!]
\centering
\resizebox{\linewidth}{!}{
\renewcommand{\arraystretch}{1.2}
\begin{tabular}{l|cc|cccc|cccc|cccc}
\toprule[1pt]
\multirow{2}{*}{\textbf{Method}} & \multirow{2}{*}{\textbf{Image Enc.}} & \multirow{2}{*}{\textbf{Text Enc.}} & \multicolumn{4}{c|}{\textbf{CUHK-PEDES}} & \multicolumn{4}{c|}{\textbf{ICFG-PEDES}} & \multicolumn{4}{c}{\textbf{RSTPReid}}  \\ \cline{4-15} 
&  &  & \textbf{R1} & \textbf{R5} & \textbf{R10} & \textbf{mAP} & \textbf{R1} & \textbf{R5} & \textbf{R10} & \textbf{mAP} & \textbf{R1} & \textbf{R5} & \textbf{R10} & \textbf{mAP} \\ 
\midrule
ViTAA \cite{wang2020vitaa} &RN50 &LSTM &55.97 &75.84  &83.52 &- &50.98 & 68.79 & 75.78 & -& - & - & -  & -  \\
SSAN \cite{ding2021semantically} & RN50 & LSTM & 61.37 & 80.15 & 86.73 & - & 54.23 & 72.63 & 79.53 & - & 43.50 & 67.80 &77.15 & - \\ 
LBUL \cite{wang2022look}& RN50& BERT & 64.04  & 82.66  & 87.22 & - & -  & - & - & - & 45.55 & 68.2  & 77.85 & -  \\
SAF \cite{li2022learning} & ViT-Base  & BERT  & 64.13  & 82.62  & 88.4  & -  & - & -  & -  & -  & -  & -  & -  & -  \\
TIPCB \cite{chen2022tipcb} & RN50 & BERT & 64.26  & 83.19  & 89.1  & -     & 54.96  & 74.72 & 81.89 & -     & -     & -     & -     & -     \\
CAIBC \cite{wang2022caibc} & RN50 & BERT & 64.43  & 82.87  & 88.37 & -     & -   & -     & -     & -     & 47.35 & 69.55 & 79.00 & -     \\
AXM-Net \cite{farooq2022axm}  & RN50  & BERT & 64.44  & 80.52  & 86.77 & 58.70 & -  & -     & -     & -     & -     & -     & -     & -     \\
LGUR \cite{shao2022learning} & DeiT-Small  & BERT  & 65.25  & 83.12  & 89.00 & - & 59.02 & 75.32 & 81.56 & - & 47.95 & 71.85 & 80.25 & -     \\
IVT \cite{shu2022see} & ViT-Base  & BERT  & 65.69  & 85.93  & 91.15 & -  & 56.04   & 73.60 & 80.22 & -  & 46.70 & 70.00 & 78.80 & -     \\
LCR²S \cite{yan2023learning}  & RN50  & TextCNN+BERT  & 67.36  & 84.19  & 89.62 & 59.20 & 57.93  & 76.08 & 82.40 & 38.21 & 54.95 & 76.65 & 84.70 & 40.92 \\
UniPT \cite{shao2023unified}  & ViT-Base & BERT   & 68.50  & 84.67  & 90.38 & -  & 60.09  & 76.19 & 82.46 & -  & 51.85 & 74.85 & 82.85 & -     \\
\midrule

\multicolumn{6}{l}{\textit{with ALBEF \cite{li2021align} backbone:}} \\
\midrule
RaSa \cite{bai2023rasa}  & CLIP-ViT & BERT-base & 76.51  & 90.29  & 94.25 & 69.38 & 65.28  & 80.40  & 85.12 & 41.29 & 66.90 & 86.50 & 91.35 & 52.31 \\
APTM \cite{yang2023towards} & Swin-B & BERT-base & 76.53  & 90.04  & 94.15 & 66.91 & 68.51 & 82.99 & 87.56 & 41.22 & 67.50 & 85.70 & 91.45 & 52.56 \\ 
\midrule
\multicolumn{6}{l}{\textit{with CLIP \cite{radford2021learning} backbone:}} \\
\midrule
Han et al. \cite{han2021text} & CLIP-RN101 & CLIP-Xformer & 64.08 & 81.73 & 88.19 & 60.08 & - & - & - & - & - & - & - & - \\
IRRA \cite{jiang2023cross} & CLIP-ViT  & CLIP-Xformer & 73.38  & 89.93  & 93.71 & 66.10 & 63.46   & 80.25 & 85.82 & 38.06 & 60.20 & 81.30 & 88.20 & 47.17 \\


\multicolumn{1}{l|}{FSRL \cite{wang2024fine}} & CLIP-ViT & \multicolumn{1}{c|}{CLIP-Xformer} & \multicolumn{1}{l}{74.65} & \multicolumn{1}{l}{89.77} & \multicolumn{1}{l}{94.03} & \multicolumn{1}{c|}{67.49} & 64.01 & 80.42 & 85.56 & \multicolumn{1}{c|}{39.64} & 60.20 & 81.40 & 88.60 & 47.38 \\

\multicolumn{1}{l|}{Propot \cite{yan2024prototypical}} & CLIP-ViT & \multicolumn{1}{c|}{CLIP-Xformer} & \multicolumn{1}{l}{74.89} & \multicolumn{1}{l}{89.90} & \multicolumn{1}{l}{94.17} & \multicolumn{1}{c|}{67.12} & 65.12 & 81.57 & 86.97 & \multicolumn{1}{c|}{\textbf{42.93}} & 61.87 & 83.63 & 89.70 & 47.82 \\

\multicolumn{1}{l|}{SAP-SAM \cite{wang2024fine}} & CLIP-ViT & \multicolumn{1}{c|}{CLIP-Xformer} & \multicolumn{1}{l}{75.05} & \multicolumn{1}{l}{89.93} & \multicolumn{1}{l}{93.73} & \multicolumn{1}{c|}{-} & 63.97 & 80.84 & 86.17 & \multicolumn{1}{c|}{-} & 62.85 & 82.65 & 89.85 & - \\

\multicolumn{1}{l|}{PLOT \cite{park2025plot}} & CLIP-ViT & \multicolumn{1}{c|}{CLIP-Xformer} & \multicolumn{1}{l}{75.28} & \multicolumn{1}{l}{90.42} & \multicolumn{1}{l}{94.12} & \multicolumn{1}{c|}{-} & 65.76 & 81.39 & 86.73 & \multicolumn{1}{c|}{-} & 61.80 & 82.85 & 89.45 & - \\

\multicolumn{1}{l|}{RDE \cite{qin2024noisy}} & CLIP-ViT & \multicolumn{1}{c|}{CLIP-Xformer} & \multicolumn{1}{l}{75.94} & \multicolumn{1}{l}{90.14} & \multicolumn{1}{l}{94.12} & \multicolumn{1}{l|}{67.56} & 67.68 & 82.47 & 87.36 & \multicolumn{1}{c|}{40.06} & 65.35 & 83.95 & 89.90 & 50.88 \\



NAM \cite{tan2024harnessing} & CLIP-ViT & CLIP-Xformer & 76.82 & 91.16 & 94.46 & 69.55 & 67.05 & 82.16 & 87.33 & 41.51 & 68.50 & 87.15 & 92.10 & 53.02 \\

Ours (1.0 M) & CLIP-ViT  & CLIP-Xformer & \underline{77.02} &\underline{91.28} &\underline{94.58} &\underline{69.65} &\underline{69.07} &\underline{83.26} &\underline{87.64} &41.91 &\underline{70.30} &\underline{88.00} &\underline{92.85}&\underline{54.89} \\

Ours (5.0 M) & CLIP-ViT  & CLIP-Xformer & \textbf{77.60} &\textbf{91.40} &\textbf{94.78} &\textbf{69.82} &\textbf{69.51} &\textbf{83.47} &\textbf{87.67} &\underline{42.30} &\textbf{71.25} &\textbf{87.25} &\textbf{92.90} &\textbf{55.43} \\
\bottomrule
\end{tabular}}
\vspace{-2mm}
\caption{Comparisons with state-of-the-art methods in the traditional evaluation settings. The best results are marked in \textbf{bold}, and the
second-best results are \underline{underlined}.}
\label{tab:sota-traditional}
\vspace{-5mm}
\end{table*}

\noindent{\bf Evaluation.} Following previous works~\cite{tan2024harnessing, qin2024noisy}, we conduct a comprehensive evaluation of our method across three challenging text-to-image person retrieval datasets: CUHK-PEDES~\cite{li2017person}, ICFG-PEDES~\cite{ding2021semantically}, and RSTPReid~\cite{zhu2021dssl}. We employ Rank-$k$ ($k$=1, 5, 10) and mean Average Precision (mAP) as evaluation metrics for all datasets.

\begin{figure}[t!]
  \centerline{\includegraphics[width=\linewidth]{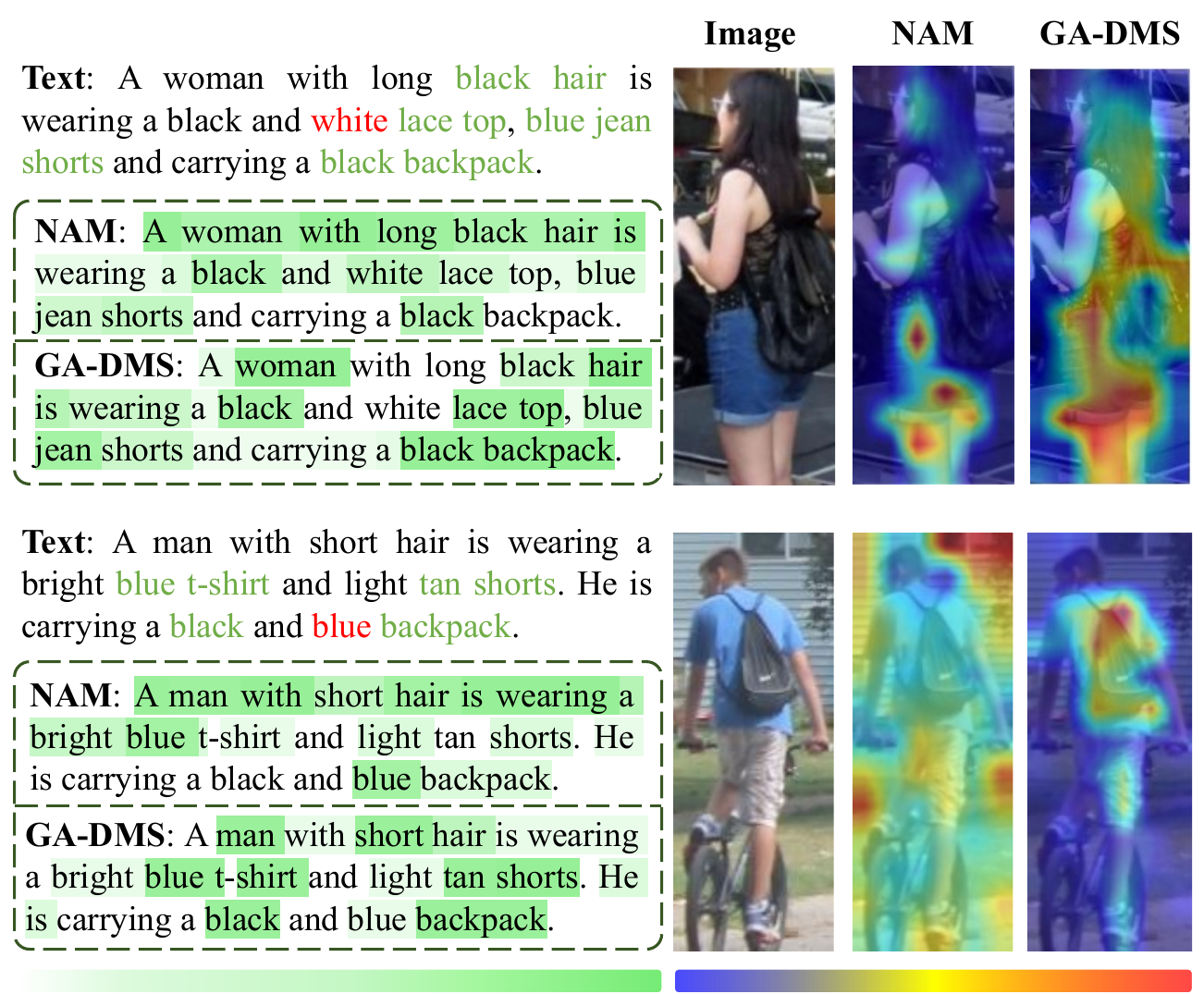}}
  \vspace{-2mm}
  \caption{Visualization of token-wise weight scores and attention maps generated by NAM~\cite{tan2024harnessing} and our GA-DMS.}
  \label{fig-visual}
  \vspace{-5mm}
\end{figure}

\begin{table}[t]
\centering
\resizebox{0.49\textwidth}{!}{%
\renewcommand{\arraystretch}{1.2}
\begin{tabular}{l|cc|cc|cc}
    \toprule
    \multirow{2}{*}{\textbf{Pre-training Dataset}} & \multicolumn{2}{c|}{\textbf{CUHK-PEDES}} & \multicolumn{2}{c|}{\textbf{ICFG-PEDES}} & \multicolumn{2}{c}{\textbf{RSTPReid}} \\ \cline{2-7} 
    & \textbf{R1} & \textbf{mAP} & \textbf{R1} & \textbf{mAP} & \textbf{R1} & \textbf{mAP} \\ 
    \midrule 
    None & 12.65 & 11.15 & 6.67 & 2.51 & 13.45 & 10.31 \\
    MALS (1.5M) & 20.47 & 18.46 & 11.71 & 4.57 & 21.50 & 16.95 \\
    LUPerson-T (0.95M) & 21.55 & 18.76 & 11.20 & 4.53 & 22.15 & 17.29 \\ 
    SYNTH-PEDES (1.0M) & 57.29  & 51.86 &\textbf{57.13}  &\textbf{31.36} & 42.20 & 32.28 \\ 
    LUPerson-MLLM (1.0M) & 56.01  & 50.34 & 37.00 & 20.21  &50.60  &37.08 \\
    \midrule
    Ours (0.1 M) & 58.95  & 52.77  & 38.18  &19.70  &47.10   &36.68  \\
    Web-Person (1.0 M) & \underline{66.26} &\underline{58.54}  &51.99 &28.81 &\underline{55.35} &\underline{40.57} \\ 
    Ours (5.0 M) & \textbf{68.34} &\textbf{60.22}  &\underline{54.64} &\underline{30.68} &\textbf{57.60} &\textbf{42.00} \\ 
    \bottomrule
\end{tabular}
}
\vspace{-2mm}
\caption{Comparisons with existing pre-training datasets in the direct transfer setting. The best results are marked in \textbf{bold}, and the
second-best results are \underline{underlined}.}
\label{tab:zero-shot}
\vspace{-5mm}
\end{table}

\subsection{Comparison with Existing Methods.} 
We evaluate GA-DMS against state-of-the-art methods on three benchmarks: CUHK-PEDES, ICFG-PEDES, and RSTPReid. As shown in Tab.~\ref{tab:sota-traditional}, our pre-trained model achieves superior performance after fine-tuning with the IRRA~\cite{jiang2023cross}, which significantly improves Rank-1 accuracy and mAP on the RSTPReid dataset by 10.10\% and 7.72\% over the baseline of IRRA.  Compared with the NAM~\cite{tan2024harnessing}, GA-DMS obtains 0.2\%, 2.02\%, and 1.8\% improvement in Rank-1 on the CUHK-PEDES, ICFG-PEDES, and RSTPReid datasets, respectively. The primary reason for this improvement is that our proposed GA-DMS framework effectively distinguishes between noise and informative tokens based on the gradient-attention similarity score. As shown in Fig.~\ref{fig-visual}, compared with NAM~\cite{tan2024harnessing}, our GA-DMS can better allocate weights to text tokens while concentrating attention on human-centric regions. This capability not only reduces the effect of noise on model training but also improves the model’s capacity to learn fine-grained semantic information. Moreover, upon scaling the WebPerson dataset from 1.0 M to 5.0 M, GA-DMS achieves new state-of-the-art Rank-1 accuracies of 77.6\%, 69.51\%, and 71.25\% across three downstream datasets.

\begin{table}[t!]
\centering
\resizebox{0.49\textwidth}{!}{%
\renewcommand{\arraystretch}{1.2}
\begin{tabular}{l|c|clclcl}
\toprule[1pt]
\multirow{3}{*}{\textbf{Pre-training Dataset}} & \multirow{3}{*}{\textbf{Source}} & \multicolumn{6}{c}{\textbf{Target}} \\ \cline{3-8} 
 &  & \multicolumn{2}{c|}{\textbf{CUHK-PEDES}} & \multicolumn{2}{c|}{\textbf{ICFG-PEDES}} & \multicolumn{2}{c}{\textbf{RSTPReid}} \\ \cline{3-8} 
 &  & \textbf{R1} & \multicolumn{1}{l|}{\textbf{mAP}} & \textbf{R1} & \multicolumn{1}{l|}{\textbf{mAP}} & \textbf{R1} & \textbf{mAP} \\ 
\midrule

\multirow{3}{*}{None} & CUHK-PEDES & \cellcolor{gray!30}{73.48} & \multicolumn{1}{l|}{\cellcolor{gray!30}{66.21}} & 43.04 & \multicolumn{1}{l|}{22.45} & 52.55 & 39.97 \\
 & ICFG-PEDES & 33.90 & \multicolumn{1}{l|}{31.65} & \cellcolor{gray!30}{63.83} & \multicolumn{1}{l|}{\cellcolor{gray!30}{38.37}} & 47.45 & 36.83 \\
 & RSTPReid & 35.25 & \multicolumn{1}{l|}{32.35} & 33.58 & \multicolumn{1}{l|}{19.58} & \cellcolor{gray!30}{60.40} & \cellcolor{gray!30}{47.70} \\ 
\midrule
 
\multirow{3}{*}{MALS(1.5 M)} & CUHK-PEDES & \cellcolor{gray!30}{73.67} & \multicolumn{1}{l|}{\cellcolor{gray!30}{65.23}} & 46.02 & \multicolumn{1}{l|}{24.06} & 55.05 & 41.29 \\
 & ICFG-PEDES & 43.11 & \multicolumn{1}{l|}{38.93} & \cellcolor{gray!30}{65.21} & \multicolumn{1}{l|}{\cellcolor{gray!30}{38.52}} & 48.45 & 37.29 \\
 & RSTPReid & 44.51 & \multicolumn{1}{l|}{39.99} & 40.78 & \multicolumn{1}{l|}{25.42} & \cellcolor{gray!30}{64.05} & \cellcolor{gray!30}{50.08} \\ 
 \midrule
 
\multirow{3}{*}{LUPerson-T(0.95 M)} & CUHK-PEDES & \cellcolor{gray!30}{74.28} & \multicolumn{1}{l|}{\cellcolor{gray!30}{66.52}} & 44.83 & \multicolumn{1}{l|}{22.72} & 54.25 & 39.26 \\
 & ICFG-PEDES & 34.66 & \multicolumn{1}{l|}{32.51} & \cellcolor{gray!30}{65.33} & \multicolumn{1}{l|}{\cellcolor{gray!30}{38.45}} & 48.30 & 38.51 \\
 & RSTPReid & 39.26 & \multicolumn{1}{l|}{34.26} & 34.95 & \multicolumn{1}{l|}{22.25} & \cellcolor{gray!30}{61.50} & \cellcolor{gray!30}{48.28} \\ 
 \midrule

\multirow{3}{*}{SYNTH-PEDES(1.0 M)} & CUHK-PEDES & \cellcolor{gray!30}{74.12} & \multicolumn{1}{l|}{\cellcolor{gray!30}{65.82}} & 57.14 & \multicolumn{1}{l|}{32.12} & 55.85 & 40.85 \\
 & ICFG-PEDES & 60.49 & \multicolumn{1}{l|}{54.61} & \cellcolor{gray!30}{66.63} & \multicolumn{1}{l|}{\cellcolor{gray!30}{39.32}} & 49.80 & 37.34 \\
 & RSTPReid & 57.75 & \multicolumn{1}{l|}{53.01} & 53.88 & \multicolumn{1}{l|}{30.88} & \cellcolor{gray!30}{66.75} & \cellcolor{gray!30}{52.18} \\ 
\midrule
 
\multirow{3}{*}{LUPerson-MLLM(1.0 M)} & CUHK-PEDES & \cellcolor{gray!30}{76.59} & \multicolumn{1}{l|}{\cellcolor{gray!30}{68.06}} & 47.17 & \multicolumn{1}{l|}{25.41} & 59.35 & 43.76 \\
 & ICFG-PEDES & 60.75 & \multicolumn{1}{l|}{54.42} & \cellcolor{gray!30}{67.18} & \multicolumn{1}{l|}{\cellcolor{gray!30}{40.27}} &55.65 & 44.05 \\
 & RSTPReid & 60.04 & \multicolumn{1}{l|}{53.85} & 46.39 & \multicolumn{1}{l|}{27.91} & \cellcolor{gray!30}{69.45} & \cellcolor{gray!30}{53.30} \\ 
\midrule
 
\multirow{3}{*}{Ours(0.1 M)} & CUHK-PEDES & \cellcolor{gray!30}{75.53} & \multicolumn{1}{l|}{\cellcolor{gray!30}{67.92}} & 47.79 & \multicolumn{1}{l|}{25.14} & 56.75 & 41.01 \\

 & ICFG-PEDES & 58.67 & \multicolumn{1}{l|}{52.66} & \cellcolor{gray!30}{66.35} & \multicolumn{1}{l|}{\cellcolor{gray!30}{39.95}} & 52.93 & 39.84 \\

 & RSTPReid & 58.49 & \multicolumn{1}{l|}{52.50} & 44.41 & \multicolumn{1}{l|}{25.98} & \cellcolor{gray!30}{65.90} & \cellcolor{gray!30}{49.28} \\ 
 \midrule
 
\multirow{3}{*}{Ours(1.0 M)} & CUHK-PEDES & \cellcolor{gray!30}{\underline{77.02}} & \multicolumn{1}{l|}{\cellcolor{gray!30}{\underline{69.65}}} & \underline{57.24} & \multicolumn{1}{l|}{\underline{32.13}} & \underline{61.10} & \underline{45.27} \\
 & ICFG-PEDES & \underline{68.16} & \multicolumn{1}{l|}{\underline{60.79}} & \cellcolor{gray!30}{\underline{69.07}} & \multicolumn{1}{l|}{\cellcolor{gray!30}{\underline{41.91}}} &\underline{59.15} & \underline{44.94} \\
 & RSTPReid & \underline{68.41} & \multicolumn{1}{l|}{\underline{61.28}} & \underline{56.13} & \multicolumn{1}{l|}{\underline{34.64}} & \cellcolor{gray!30}{\underline{70.30}} & \cellcolor{gray!30}{\underline{54.89}} \\ 
\midrule

\multirow{3}{*}{Ours(5.0 M)} & CUHK-PEDES & \cellcolor{gray!30}{\textbf{77.60}} & \multicolumn{1}{l|}{\cellcolor{gray!30}{\textbf{69.82}}} & \textbf{58.91} & \multicolumn{1}{l|}{\textbf{33.70}} & \textbf{61.80} & \textbf{46.81} \\
 & ICFG-PEDES & \textbf{69.83} & \multicolumn{1}{l|}{\textbf{62.06}} & \cellcolor{gray!30}{\textbf{69.52}} & \multicolumn{1}{l|}{\cellcolor{gray!30}{\textbf{42.30}}} &\textbf{60.05} & \textbf{45.46} \\
 & RSTPReid & \textbf{69.19} & \multicolumn{1}{l|}{\textbf{62.00}} & \textbf{57.13} & \multicolumn{1}{l|}{\textbf{35.76}} & \cellcolor{gray!30}{\textbf{71.25}} & \cellcolor{gray!30}{\textbf{55.43}} \\ 
 \bottomrule
\end{tabular}
}
\vspace{-2mm}
\caption{Comparisons with existing pre-training datasets in the fine-tuning setting. The best results are marked in \textbf{bold}, and the second-best results are \underline{underlined}. \colorbox{gray!30}{Gray} indicates that the source and target are homologous.}
\label{tab:sota-finetune}
\vspace{-5mm}
\end{table}

\subsection{Comparison with Existing Datasets.} We conduct comprehensive comparisons between our WebPerson dataset and four existing large-scale pre-training datasets: MALS~\cite{yang2023towards}, LUPerson-T~\cite{shao2023unified}, SYNTH-PEDES~\cite{zuo2024plip}, and LUPerson-MLLM~\cite{tan2024harnessing}. MALS consists of 1.5 million synthetic images generated using commercial diffusion models, with textual descriptions automatically produced by BLIP~\cite{li2022blip}. LUPerson-T includes 0.95 million images, each enhanced by one of 456 templates to maximize caption diversity. SYNTH-PEDES provides 4.8 million images, each annotated with an average of 2.53 textual descriptions, generated through a hybrid architecture that combines a ResNet101-FPN~\cite{he2016deep} visual encoder with a GPT-2~\cite{radford2019language} text generator for detailed person attribute modeling. Notably, LUPerson-MLLM utilizes two multimodal large language models for caption generation, supplemented by 46 ChatGPT-optimized templates obtained through iterative dialogues to enhance linguistic variation. This dataset comprises 1.0 million images, each paired with two MLLM-generated captions.

Tab.~\ref{tab:zero-shot} presents comparative results under a direct transfer setting, where models pre-trained on Web-Person exhibit superior cross-dataset generalization across three benchmarks. Specifically, under the comparable 1M dataset, our constructed WebPerson dataset demonstrates superior performance on CUHK-PEDES and RSTPReid, and shows suboptimal performance on ICFG-PEDES. Notably, the WebPerson dataset demonstrates comparable performance to the full-scale LUPerson-MLLM even when trained on a mere 0.1M samples. These experimental results demonstrate that our proposed WebPerson dataset exhibits strong robustness and can learn person representations with enhanced transferability.

As shown in Tab.~\ref{tab:sota-finetune}, we also evaluate the fine-tuning performance following LUPerson-MLLM~\cite{tan2024harnessing}, utilizing the IRRA with models pretrained on different datasets. Results indicate that WebPerson pretraining yields state-of-the-art performance across both in-domain and cross-domain scenarios. At the 1M data scale, WebPerson achieves consistent improvements over LUPerson-MLLM, with Rank-1 accuracy gains of 0.43\%, 1.89\%, and 0.85\% on CUHK-PEDES, ICFG-PEDES, and RSTPReid respectively. The cross-domain evaluations reveal particularly significant performance enhancements, highlighting WebPerson's exceptional representation transferability through fine-tuning.

\subsection{Ablation Study}

\begin{table}[t!]

\centering
\setlength{\tabcolsep}{4pt} 
\resizebox{0.49\textwidth}{!}{%
\renewcommand{\arraystretch}{1.2}
\begin{tabular}{cc|cc|cc|cc|cc}
\toprule
\multicolumn{2}{c|}{\textbf{Masking Method}} & \multicolumn{2}{c|}{\textbf{Components}} & \multicolumn{2}{c|}{\textbf{CUHK-PEDES}} & \multicolumn{2}{c|}{\textbf{ICFG-PEDES}} & \multicolumn{2}{c}{\textbf{RSTPReid}} \\
\midrule
\textbf{CSS} & \textbf{GASS} & \textbf{SDM} & \textbf{MTP} & \textbf{R1} & \textbf{mAP} & \textbf{R1} & \textbf{mAP} & \textbf{R1} & \textbf{mAP} \\
\midrule
\xmark &\xmark &\xmark &\xmark  & 56.75 & 50.42 & 34.63 & 17.59 & 45.50 & 34.51 \\
\midrule
\checkmark & \xmark &\xmark & \checkmark & 56.35 & 50.21 & 34.72 & 17.66 & 44.60 & 33.28 \\
\checkmark &\xmark & \checkmark &\xmark & 63.29 & 57.42 & 43.39 & 24.12 & 51.95 & 39.41 \\
\checkmark &\xmark & \checkmark & \checkmark & 62.74 & 57.01 & 42.96 & 23.88 & 50.80 & 38.91 \\
\midrule
\xmark &\checkmark &\xmark & \checkmark& 57.29 & 52.28 & 36.24 & 18.96 & 47.90 & 35.97 \\
\xmark &\checkmark & \checkmark &\xmark & \underline{63.87} & \underline{57.56} & \underline{44.02} & \underline{24.18} & \underline{52.30} & \underline{39.61} \\
\xmark &\checkmark & \checkmark & \checkmark & \textbf{64.25} & \textbf{58.27} & \textbf{44.39} & \textbf{24.67} & \textbf{52.70} & \textbf{40.12} \\
\bottomrule
\end{tabular}
}
\vspace{-2mm}
\caption{Ablation on different components and masking methods. CSS: Cosine Similarity Score. GASS: Gradient-Attention Similarity Score. SDM: Similarity Distribution Matching. MTP: Masked Token Prediction.}
\label{tab:ablation_components}
\vspace{-5mm}
\end{table}
\noindent{\bf Ablation on Different Components and Masking Methods.}
To substantiate the efficacy of various components and the effectiveness of our proposed Gradient-Attention Similarity Score~(GASS), we perform a comprehensive ablation study with a 0.5M data sample from our WebPerson dataset. As shown in Tab.~\ref{tab:ablation_components}, the integrating Masked Token Prediction (MTP) with GASS improves performance across all evaluation metrics, as predicting semantically rich tokens enhances fine-grained learning. The Similarity Distribution Matching (SDM) component alone enhances image-text alignment by replacing noisy tokens with learnable embeddings, achieving Rank-1 accuracy gains of 7.12\%, 9.39\%, and 6.8\% on CUHK-PEDES, ICFG-PEDES, and RSTPReid respectively. By combining MTP with SDM, we observe enhancements across all metrics, further substantiating the efficacy of the components within our method.

When comparing Cosine Similarity Score (CSS) with Gradient-Attention Similarity Score (GASS), GASS consistently exhibits superior performance. This advantage primarily stems from GASS's capacity to precisely weight textual tokens during training by incorporating gradient and attention information. As illustrated in Fig.~\ref{fig-visual}, our method accurately allocates weights to noise textual tokens (\textit{e.g.}, "white lace top"), thereby effectively mitigating the influence of noise on the model's representation learning.

\begin{figure}[t!]
\begin{subfigure}[b]{\linewidth}
\centering
\includegraphics[width=1.0\linewidth]{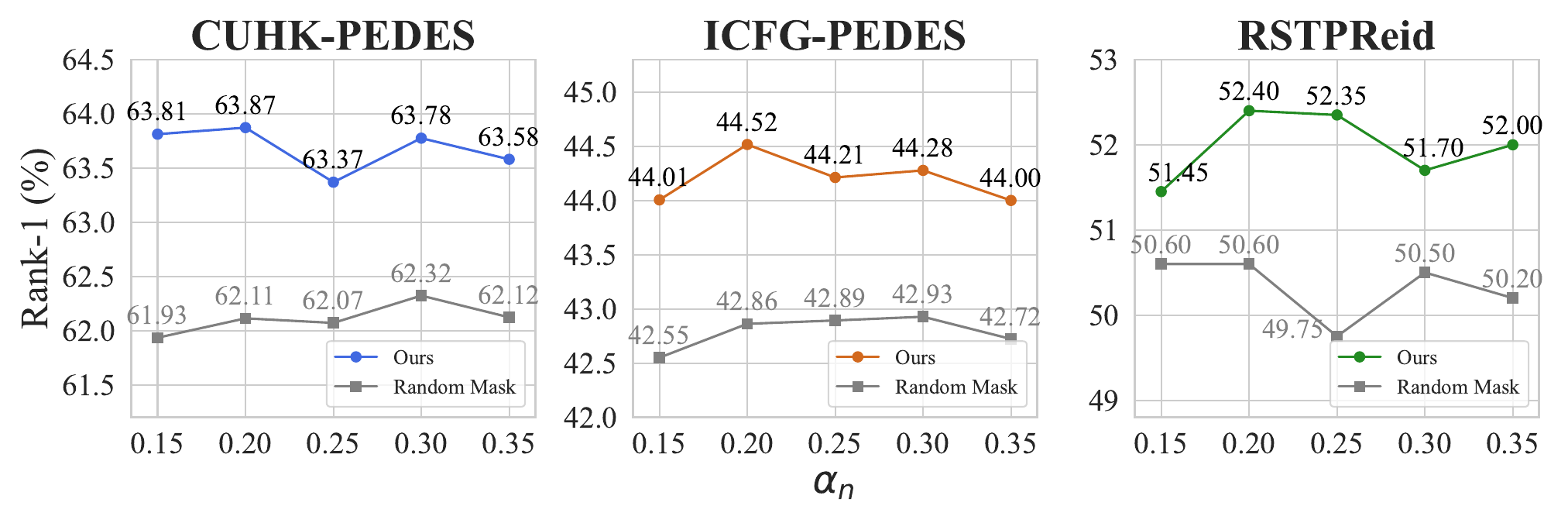}
\vspace{-7mm}
\subcaption{\small Ablation on $\alpha_{n}$ for masking noise tokens}
\label{fig:NAM}
\end{subfigure}
\hfill
\begin{subfigure}[b]{\linewidth}
\centering
\includegraphics[width=1.0\linewidth]{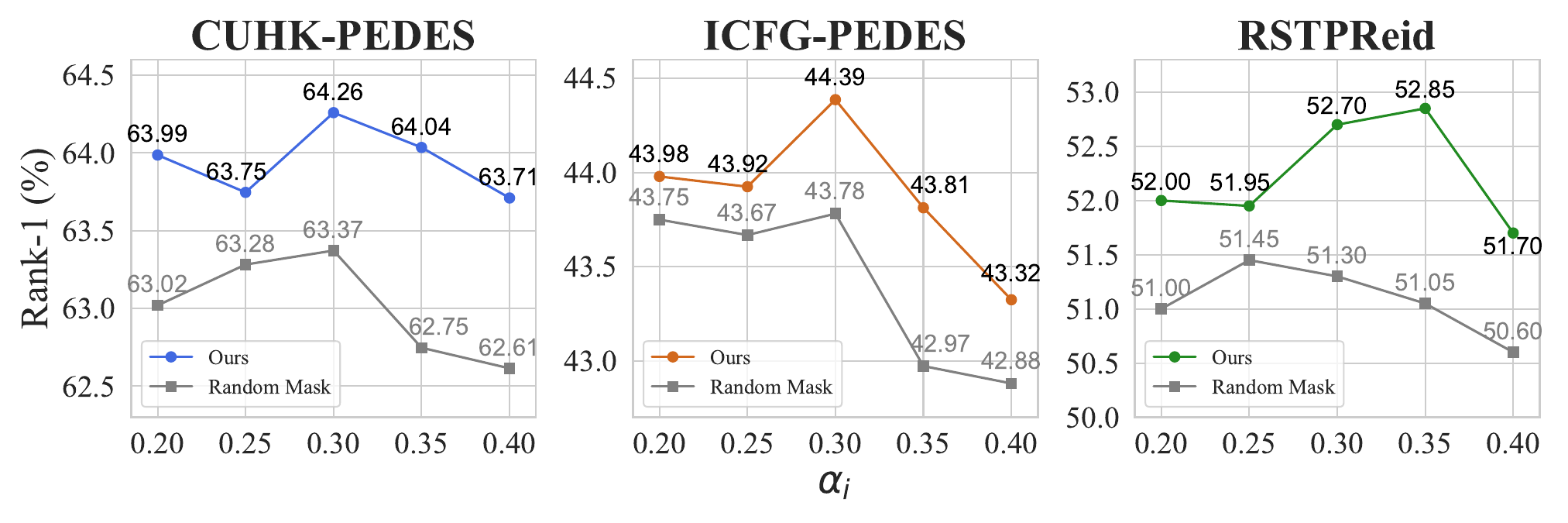}
\vspace{-7mm}
\subcaption{\small Ablation on $\alpha_{i}$ for masking informative tokens}
\label{fig:MLM}
\end{subfigure}
\vspace{-7mm}
\caption{Ablation experiment results for $\alpha_{n}$ and $\alpha_{i}$, which can directly influence the upper limit of the masking probability for noise and informative tokens.}
\vspace{-3mm}
\label{fig:Ratio}
\end{figure}

\noindent{\bf Ablation on \textbf{$\alpha_{n}$} and \textbf{$\alpha_{i}$}.} 
In this work, our dual-masking synergetic learning method dynamically masks textual tokens according to gradient-attention similarity scores. We introduce parameters $\alpha_{n}$ and $\alpha_{i}$ to regulate the maximum masking probabilities for noise and informative tokens. Fig.~\ref{fig:Ratio} presents an ablation study on $\alpha_{n}$ and $\alpha_{i}$ to determine the optimal settings. For enhanced performance on three downstream datasets, we set $\alpha_{n}=0.2$ and $\alpha_{i}=0.3$. Additionally, our method consistently outperforms random masking baselines, confirming its effectiveness.

\begin{figure}[t]
  \centerline{\includegraphics[width=1.0\linewidth]{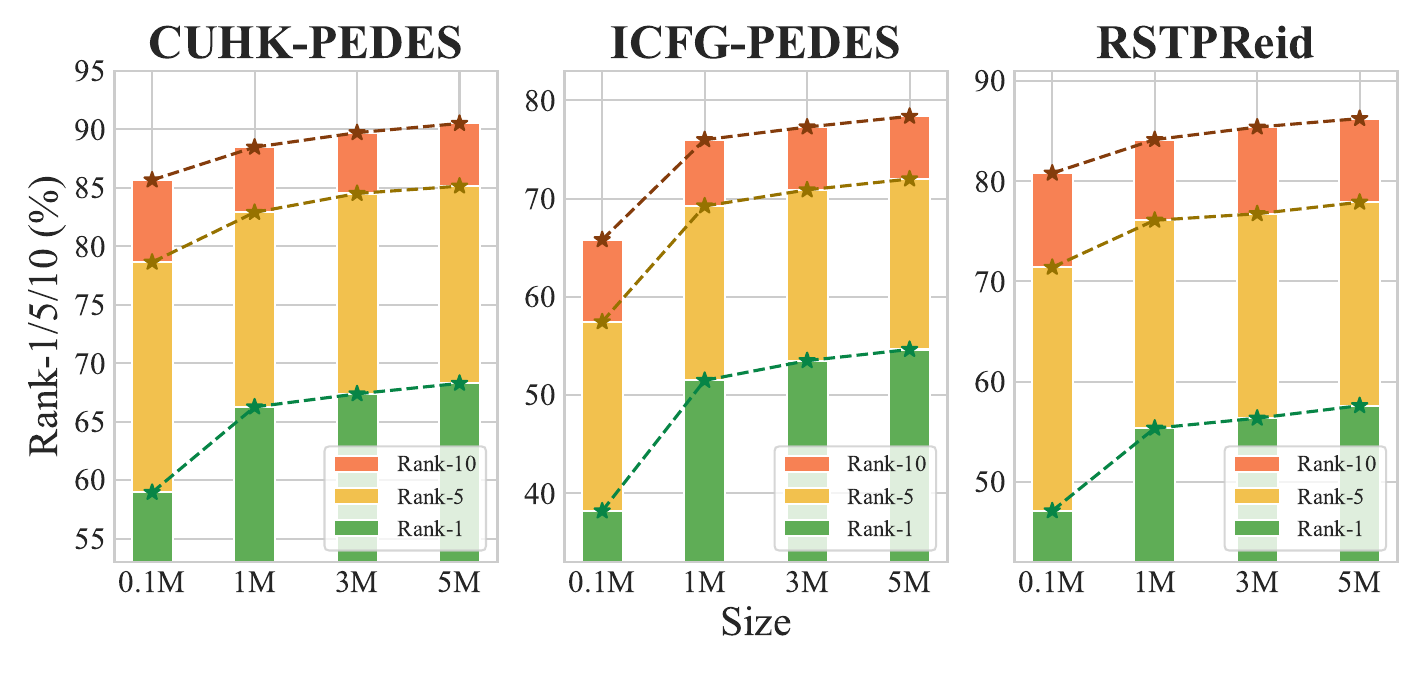}}
  \vspace{-5mm}
  \caption{Data scaling analysis of WebPerson dataset.The performance of our GA-DMS method in direct transfer settings.}
  \label{fig:size}
  \vspace{-5mm}
\end{figure}

\noindent{\bf Data Scaling Analysis.}
To explore the impact of pretraining data scale on person representation learning, we systematically augmented the dataset size from 0.1M to 1M, 3M, and 5M samples for pertaining. Fig.~\ref{fig:size} illustrates the direct transfer performance evaluation across three benchmarks at different data scales. The outcomes consistently reveal performance enhancements as the data volume increases. At the maximum scale of 5.0M samples, the model demonstrates Rank-1 accuracy improvements of 9.39\%, 16.46\%, and 10.50\% across the three benchmarks in comparison to the 0.1M baseline, indicating a clear upward trajectory. These findings conclusively demonstrate that scaling high-quality pretraining data substantially enhances text-based person retrieval capability.

\section*{Conclusion}
In this paper, we enhance CLIP for person representation learning by synergistically improving data acquisition and model architecture. First, we devise a noise-resistant data construction pipeline that leverages the in-context learning capabilities of MLLMs for automatic filtering and captioning of web-crawled images. This results in the WebPerson dataset, which comprises 5M high-quality person-centric image-text pairs. Second, we propose the GA-DMS framework, which improves cross-modal alignment by masking semantically irrelevant tokens based on a gradient-attention similarity score. Concurrently, we implement masked token prediction objectives that force the model to reconstruct informative text tokens, facilitating discriminative fine-grained feature learning for visual-semantic correspondence. Comprehensive experiments demonstrate that GA-DMS achieves state-of-the-art performance in several downstream datasets. We hope our work provides insights for the person representation learning task.
\section*{Limitations}
In this work, we demonstrate the exceptional text-based person retrieval performance of the person-centric dataset constructed solely from internet images. Limited by computational resources, this paper constructs a 5M-scale WebPerson dataset, with further scaling left for community exploration.

\section*{Acknowledgement}
This work was supported in part by the National Natural Science Foundation of China under Grant 62373086,  by the Liaoning Province Applied Basic Research Program (2025JH2/101330131), by the Guangdong Basic and Applied Basic Research under 2023A1515140014, and by the State Key Laboratory of Robotics under Grant 2024-O14.
\section*{Ethics Statement}
We abide by the ACL Code of Ethics. The data resources used in this study are publicly available.

\bibliography{custom}

\clearpage

\appendix
\section{Appendix}
\subsection{Detail Experimental Settings}
\label{Appendix hyper}
We present the settings used in the training GA-DMS in Tab.~\ref{tab:hyperparams}.

\begin{table}[h]
    \centering
    \resizebox{0.8\linewidth}{!}{
     \begin{tabular}{l|c}
        \toprule Hyperparameter & Value \\
        \midrule
        Temperature & $0.02$ \\
        Loss weight $\beta$ & $0.4$ \\ 
        Multiple scales $\mathcal{C}$ & [1,2]\\
        Adam $\beta_{1}$ & $0.9$ \\
        Adam $\beta_{2}$ & $0.999$ \\
        Adam $\epsilon$  & $10^{-3}$ \\
        Warm-up epochs & 5\\
        Weight decay & $4\times 10^{-5}$ \\
        Batch size & $512$ \\
        Learning rate & $10^{-4}$ \\
        Learning rate scheduler & CosineAnnealingLR \\
        Training epochs & $30$  \\
        
        GPU & $8 \times $A100(80G) \\
        \bottomrule
    \end{tabular}
    }
\centering
\caption{Hyperparameters used for GA-DMS pre-training.}
\label{tab:hyperparams}
\vspace{-5mm}
\end{table}

\subsection{Detail Instruction Prompt}
\label{Appendix prompt}
The prompt used to input Qwen2.5-72B-Instruct~\cite{yang2024qwen2} for the generation of structured templates is as follows:

\emph{\color{black!65} First, identify the words in the title that describe pedestrian attributes, such as tops, pants, footwear, head features, accessories, age, gender, actions, etc. Then replace these words with cross-identity generic terms like ‘colored top’, ‘colored bottom’,‘hairstyle’ etc. Complete examples are as follows:}

\emph{\color{black!65}"A man wearing a orange jersey with yellow stripes, a pair of black shorts and a pair of green shoes." $\rightarrow$” A [man] wearing a [color top] with [color pattern], a pair of [colored bottom] and a pair of [colored shoes].” }

\emph{\color{black!65}“This lady is wearing glasses, and she has her hair in a yellow ponytail. She is wearing a striped shirt and is carrying a bag over her right shoulder." $\rightarrow$ “This [person] is wearing an [accessory], and [he/she] has a [colored hairstyle]. [He/She] is wearing a patterned top and is carrying an object over [his/her] [body part].”}

\emph{\color{black!65}“A women is wearing a light colored sweater and black pants. She has long dark hair in a pony tail. " $\rightarrow$ "A [person] is wearing a [colored top] and [colored bottom]. [He/She] has long [colored hair] in a [hairstyle]."}

\emph{\color{black!65}Do not add any extra features not included in the original description. Output only the final description without any explanation.}

The prompt used for inputting Qwen2.5-VL-Instruct~\cite{bai2025qwen2} to generate pedestrian descriptions is as follows:


\emph{\color{black!65}"Please generate a concise caption for the pedestrian image based on the following principles:}

\emph{\color{black!65}Core Subject Focus: Only describe the dominant pedestrian elements in the frame (e.g., gender, clothing, footwear, head features, accessories, actions),focusing on the color of each part."}

\emph{\color{black!65}Description restriction: 1.Use vague color terms (e.g., dark, light) only when the color is uncertain. 2.Use generic terms like "top" or "bottom" only when the clothing type is unclear, otherwise, use specific terms like "shirt" or "shorts."}

\emph{\color{black!65}Background Suppression Rule: Do not mention background information or abstract atmospheres (e.g., cozy).}

\emph{\color{black!65}Certainty Principle: Only output visually confirmed details — omit descriptions of unclear/low-resolution areas. Invisible elements do not need be described in the sentence(e.g., items are not visible). Avoid speculative terms ("possibly", "seems", "appears to be"), do not interpret potential relationships (e.g., inferring identity or emotions), and exclude artistic style critiques (e.g., "impressionist style").}

\emph{\color{black!65}Sentence Structure Reference: "<Structured Template>",First output the most significant pedestrian elements, the sentence length is less than <sequence length> English words. 
Use common words and phrasing from social media or daily life, ensuring correct grammar and logic. Provide only the caption sentence without any additional output."}


\subsection{The Influence of Layers.}

\begin{figure}[t!]
\centerline{\includegraphics[width=1.0\linewidth]{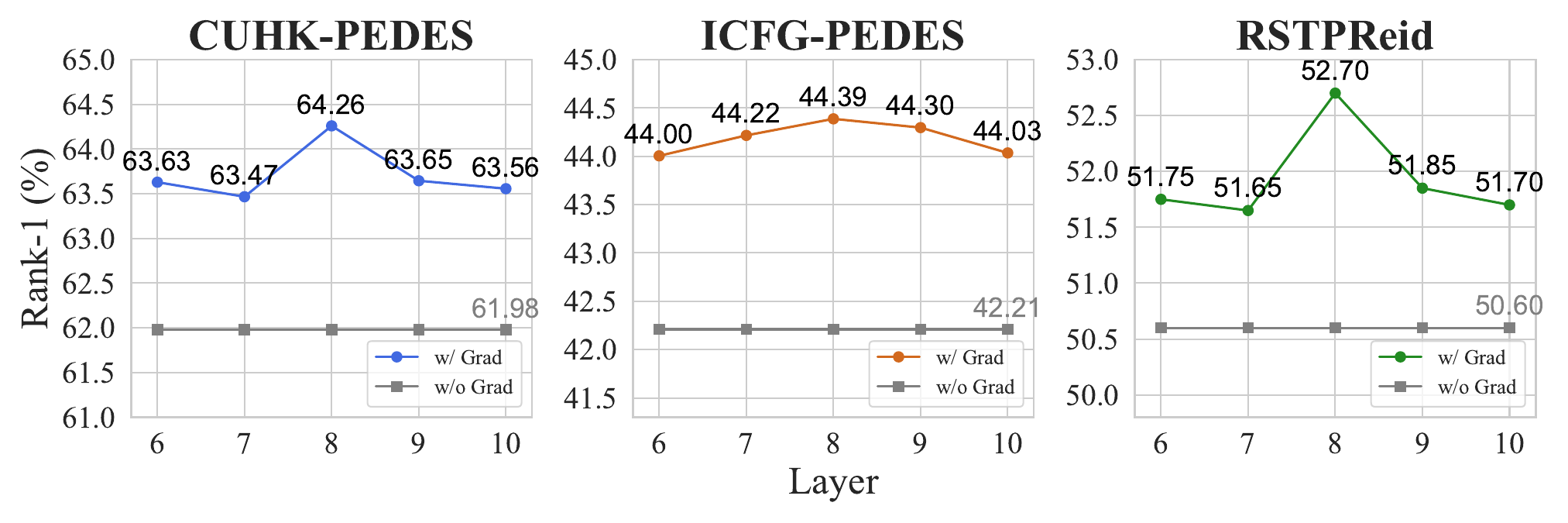}}
\vspace{-2mm}
  \caption{Results of different layers to compute $\mathbb{S}$. The encoders contain 12 layers in total.}
  \label{fig:layer}
  \vspace{-5mm}
\end{figure}

We calculate the Gradient-Attention Similarity Score (GASS) between each text token and the image using the final $L$ layers of the text encoder. This study examines how the number of layers involved in gradient-based similarity computation influences performance. As depicted in Fig.~\ref{fig:layer}, the model consistently outperforms the baseline, which lacks gradient-based masking, across all tested layer depths. Notably, employing the last $8$ layers of the text encoder achieves the highest overall performance, underscoring their effectiveness in optimizing masking outcomes.

\begin{table*}[t!]
\centering

\label{table:dataset}
\resizebox{0.95\linewidth}{!}{
\begin{tabular}{c|c|c|c|c|c|c}
\hline 
Datasets                  &Year &\#Images &\#Descriptions &Data Source &\#Vocabulary Size & Label Method  \\
\hline 
CUHK-PEDES~\cite{li2017person}&2017 &40,206 &80,412 &Market, Duke, etc. &12,517 &Manual \\
LPW~\cite{song2018region}            &2018 &592,438&-      &Surveillance Video                 &-  &Manual+Detector+NN\\
MSMT-17~\cite{MSMT17}     &2018 &126,441&-      &Manual Collection   &-  &FasterRCNN\\
RSTPReid~\cite{zhu2021dssl}  &2021 &20,505 &41,010  &MSMT-17             &6,331 &Manual \\
ICFG-PEDES~\cite{ding2021semantically}    &2021 &54,522 &54,522 &MSMT-17             &5,848 &Manual \\
LUPerson~\cite{fu2021unsupervised}       &2021 &4,180,243&-    &YouTube             &-  &YOLOv5 \\
LUPerson-NL~\cite{fu2022large}  &2022 &10,683,716&-   &YouTube             &-  &FairMOT\\
MALS~\cite{yang2023towards}          &2023 &1,510,330&1,510,330&Automatic Synthesis&4,772 &ImaginAIry\\
LuPerson-T~\cite{shao2023unified}&2023&957,606 &1,277,991   &LUPerson       &459 &CLIP\\
Luperson-MLLM~\cite{tan2024harnessing}&2024&1,020,022   &2,037,239   &LUPerson &39,566  &MLLM\\
SYNTH-PEDES~\cite{zuo2024plip}&2024  &4,791,771   &12,138,157  &LUPerson-NL\& LPW  &8,598  &SPAC\\
\hline
\cellcolor{gray!20}\textbf{WebPerson}&\cellcolor{gray!20}2025&\cellcolor{gray!20}5,002,723&\cellcolor{gray!20}10,005,446&\cellcolor{gray!20}COYO-700M &\cellcolor{gray!20}96,623 &\cellcolor{gray!20}MLLM \\
\hline 
\end{tabular}}
\vspace{-2mm}
\caption{Statistical comparison of different datasets. WebPerson stands as the largest automatically-generated text-described person dataset, offering inherent scalability without manual annotation requirements.}
\label{dataset}
\vspace{-3mm}
\end{table*}

\subsection{Dataset analysis}
\begin{figure}[t!]
\centering
\includegraphics[width=\linewidth]{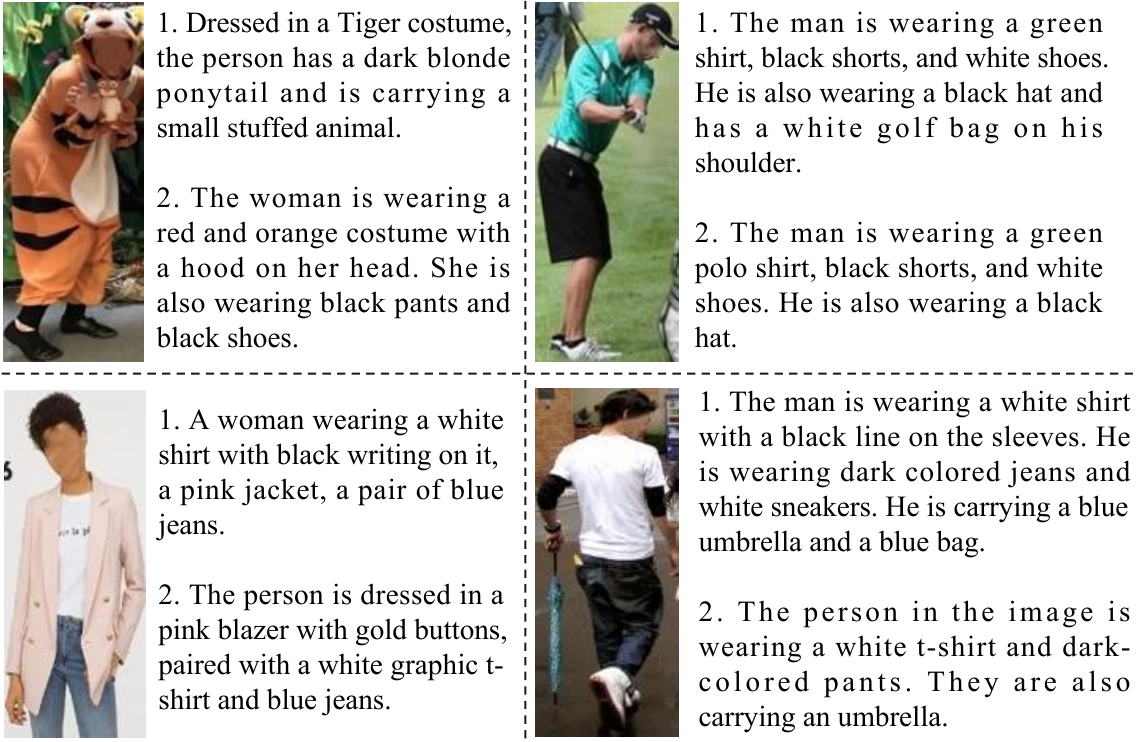}
\caption{Visualization of some examples in our WebPerson dataset.}

\label{visual-dataset}
\vspace{-3mm}
\end{figure}

Current text-based person retrieval datasets predominantly consist of manually annotated pedestrian images from re-identification benchmarks, fundamentally limited in scale and diversity by the substantial costs of human annotation. While generative methods have shown promise for dataset augmentation, they fail to achieve the necessary scale and fidelity for practical deployment. The emergence of Multimodal Large Language Models (MLLMs) and the availability of web-scale image resources now enable a new paradigm for automated dataset construction. Our WebPerson dataset leverages novel image filtering and text generation techniques to create a comprehensive pedestrian image library with accurate textual descriptions across diverse scenarios. Compared to existing datasets, WebPerson offers three key advantages:

\noindent{\bf High-quality} WebPerson surpasses existing datasets containing single-style synthetic images or low-quality surveillance footage by providing superior texture details and diverse scene variations. Our rigorous image filtering pipeline ensures exceptional visual fidelity, while the MLLM-powered text generation framework produces highly accurate and detailed descriptions. Fig.~\ref{visual-dataset} showcases representative examples demonstrating precise textual characterization of pedestrian attributes.

\noindent{\bf Diversity} Sourced from web data, WebPerson exhibits rich variations in images, including but not limited to scene diversity, viewpoint changes, occlusions, clothing variations, and body poses. Our caption generation strategy further ensures corresponding textual descriptions maintain sufficient diversity. This dual-modality diversity enables WebPerson to serve as an effective training corpus for developing robust models that generalize well to novel and unseen data across visual tasks, language tasks, and vision-language tasks.

\noindent{\bf Large-scale} As illustrated in Tab.~\ref{dataset}, we compare the attributes of WebPerson with other prominent person datasets. WebPerson emerges as the most extensive real-world dataset, featuring high-quality image-text pairs, encompassing 5 million images and 10 million textual descriptions. Moreover, our efficient data collection and caption generation strategies enable seamless scalability in data volume.

\subsection{Broader Impact}
\label{broad_impact}
This work introduces a novel pedestrian representation learning framework that achieves fine-grained cross-modal alignment through gradient-based token-wise similarity scoring while effectively suppressing noise interference. Complementing this framework, we construct WebPerson, a large-scale human-centric dataset with diverse web-sourced image-text pairs. Together, these contributions demonstrate robust performance in human-oriented applications, including intelligent surveillance and autonomous retail systems.

\end{document}